# Markov Random Field Segmentation of Brain MR Images

Karsten Held[1,2], Elena Rota Kops[1], J. Bernd Krause[1], William M. Wells III[3]
Ron Kikinis[3], Hans-Wilhelm Müller-Gärtner[1]

[1]Institut für Medizin, Forschungszentrum Jülich, D-52425 Jülich,
[2]Institut für Theoretische Physik, Universität Augsburg, D-86135 Augsburg,
[3]Harvard Medical School and Brigham and Women's Hospital, Department of Radiology, 75 Francis St., Boston, MA 02115

Correspondence to:

Dr. E. Rota Kops
Institut für Medizin
Forschungszentrum Jülich GmbH
D - 52425 Jülich, Germany

Fax Nr. +49 - 2461 - 61-2770
e-mail: e.rota.kops@fz-juelich.de

*Abstract*

We describe a fully-automatic 3D-segmentation technique for brain MR images. By means of Markov random fields the segmentation algorithm captures three features that are of special importance for MR images: non-parametric distributions of tissue intensities, neighborhood correlations and signal inhomogeneities. Detailed simulations and real MR images demonstrate the performance of the segmentation algorithm. In particular the impact of noise, inhomogeneity, smoothing and structure thickness is analyzed quantitatively. Even single-echo MR images are well classified into gray matter, white matter, cerebrospinal fluid, scalp-bone, and background. A *simulated annealing* and an *iterated conditional modes* implementation are presented.

I. INTRODUCTION

Excellent soft-tissue contrast and high spatial resolution make magnetic resonance imaging *the* method for anatomical imaging in brain research. Segmentation of the MR image into different tissues, i.e. gray matter (GM), white matter (WM), cerebrospinal fluid (CSF), scalp-bone and other non-brain tissues (SB) and background (BG)[1], is an important prerequisite for

- 3D visualization and peeling off non-brain structures,
- quantitative analysis of brain morphometry,
- matching MR onto functional images [1], [2],
- partial volume correction [3].

---

[1] For clinical applications specific abnormal tissues can be added.

Though several segmentation techniques are present in the literature, the fully-automatic segmentation of MR images remains difficult (for a review see [4]), mainly due to the somewhat noisy MR data, caused by time and equipment limitations. Another persistent difficulty is the spatial inhomogeneity of the MR signal. Some promising algorithms exist, but most of them require multi-echo images and some pre- and post-processing to improve segmentation. To this end non-linear filters [5], [6], morphological [7] and connectivity [8] operations and even snake models [9] are discussed in the literature. Besides additional effort, these processes frequently tend to suppress fine details of high-resolution images. The improvement undoubtedly obtained by the use of multi echo (high-resolution) MR images imposes the penalty of long acquisition times, which are unacceptable for some applications.

In order to address these difficulties, we have developed a new Markov random field (MRF) segmentation algorithm based on an *adaptive segmentation* algorithm described by Wells et al. [10]. The new approach uses MRF as a convenient means for introducing context or dependence among neighboring voxels. It incorporates the following important characteristics:

1. non-parametric distribution of tissue intensities are described by Parzen-window statistics [11],
2. neighborhood tissue correlations are taken into account by MRF to manage the noisy data,
3. signal inhomogeneities are also modeled by *a priori* MRF.

Several methods for addressing these issues can be found in the literature, but the algorithm presented here is the first that addresses all three simultaneously.

Wells et al. [10] use non-parametric, Parzen-window statistics [11] and adapt a bias field to the inhomogeneities but do not regard neighborhood dependencies for the tissue segmentation. Additional filtering and connectivity operations are also used.

Geman and Geman [12] were the first to apply the methods of statistical mechanics[2] to image segmentation. They use an *a priori* probability model for neighboring voxels and some additional, hidden edge elements. But they do not take account of non-parametric intensity distributions and the inhomogeneities that are important for MR images.

Pappas [16] adaptive K-means clustering algorithm uses neighborhood dependencies but only parametric Gaussian intensity distributions. Inhomogeneities are spatially adapted by adjusting the Gaussian means. In medical imaging this algorithm was applied to CT images [17]. Related MRF clustering algorithms without inhomogeneity correction were also employed for MR images [18], [19], [20].

In MATERIALS AND METHODS we explain the MRF segmentation, i.e. the statistical model (capturing the neighborhood MRF, the inhomogeneity MRF and the Parzen-window intensity distribution) and the optimization (by *simulated annealing* (SA) or *iterated conditional modes* (ICM)). The acquisition and simulation of MR images is described thereafter. The segmentation of these images and a quantitative analysis of the impact of noise, inhomogeneity, smoothing and structure thickness follow in section III. In section IV we discuss the advantages and disadvantages of the SA and two ICM algorithms, differing in the inhomogeneity correction and compare the MRF segmentation to the *adaptive segmentation* [10].

---

[2] The same methods are employed in statistical mechanics for microscopic models, on a $O(10^{-6})$ finer scale, to calculate the $T_1$ and $T_2$ relaxation times [13], [14], [15] used in MR imaging to discriminate between tissues.

## II. MATERIALS AND METHODS

*A. Markov Random Field Segmentation*

A natural way of incorporating spatial correlations into a segmentation process is to use Markov random fields [12], [16], [21], [22] as *a priori* models. The MRF is a stochastic process that specifies the *local* characteristics of an image and is combined with the given data to reconstruct the true image. The MRF of prior contextual information is a powerful method for modeling spatial continuity and other features, and even simple modeling of this type can provide useful information for the segmentation process. The MRF itself is a *conditional* probability model, where the probability of a voxel depends on its neighborhood. It is equivalent to a Gibbs *joint* probability distribution [12] determined by an energy function. This energy function is a more convenient and natural mechanism for modeling contextual information than the local *conditional* probabilities of the MRF. The MRF on the other hand is the appropriate method to sample the probability distribution.

In the following we denote the observed MR echo intensities as **z**, a vector containing an individual intensity $z_i$ for every voxel *i* in the 3D MR image. The segmentation task is to classify the tissue of every voxel *i*, i.e. to determine the segmentation vector **x** with discrete values $x_i \in \{BG, WM, GM, CSF, SB\}$. To model intensity inhomogeneities an additional vector **y** has to be calculated. By multiplying $y_i$ with the echo intensity $z_i$, the inhomogeneity corrected MR echo is obtained [10].

As pointed out in the introduction, the MR segmentation algorithm includes non-parametric statistics, neighborhood correlations and signal inhomogeneities. A MRF *a priori* probability p(**x**) for the segmented image is used to model the spatial correlations within the image. A smooth inhomogeneity field **y** is provided by a restrictive *a priori* MRF distribution p(**y**). Given the *a priori* probabilities for the tissue $x_i$ and the inhomogeneity $y_i$ at a voxel *i*, the conditional probability of the observed echo intensity $z_i$ is calculated by a Parzen-window distribution $p(z_i|x_i,y_i)$.

For given MR intensities **z**, Bayes rule is used to calculate the *a posteriori* probability of the segmentation **x** and the inhomogeneity **y**

$$p(\mathbf{x},\mathbf{y}|\mathbf{z}) \propto p(\mathbf{z}|\mathbf{x},\mathbf{y})\, p(\mathbf{x})\, p(\mathbf{y}). \qquad (1)$$

This probability is maximized either by *simulated annealing* or *iterated conditional modes*. The corresponding segmentation **x** is called the *maximum a posteriori* (MAP) estimator of the true image. The three terms on the right side of eqn. (1) are now explained.

Neighborhood MRF

The discrete Fourier transformation to reconstruct MR image has the effect that neighboring voxels contribute to echo intensities by a *sinc* function. Partial volume effects additionally smooth fine anatomical structures. These effects and especially the anatomic arrangement of the tissues lead to spatial correlations in the MR intensities. These correlations are exploited to manage with the somewhat noisy MR data and, hence, to improve the segmentation.

Balancing computing time versus *a priori* information the first order neighborhood of 6 nearest neighbors was chosen. Including dependencies between more voxels will dramatically increase the computing time, e.g. by a factor of 26/6 if all third order neighbors of a voxel are included. On the contrary, information gain is small since strong dependencies between distant voxels cannot model the complex, convoluted boundaries between brain tissues. With a neighborhood system of 6 nearest neighbors in 3D we employ the probability distribution

$$p(\mathbf{x}) \propto e^{-E(\mathbf{x})} \qquad (2)$$

with the Gibbs energy

$$E(\mathbf{x}) = \sum_{<i,j>} e_{x_i x_j}, \tag{3}$$

where $<i,j>$ denotes the sum over all voxels $i$ and their 6 nearest, first order neighbors $j$. The local potential $e_{x_i x_j}$ is the most general potential between two voxels. It must be minimal for neighboring voxels of the same tissue ($x_i = x_j$) to prefer equal segmentation at neighboring voxels, i.e. to incorporate the spatial correlations within the MR image.

The additional *a priori* information that the scalp-bone tissue is not connected with cerebral tissues is used by setting a high potential $e_{x_i x_j}$ for $x_i$ = SB and $x_j \in$ {GM,WM,CSF}, or vice versa. This potential facilitates the correct classification of scalp-bone even with strongly varying intensities.

Inhomogeneity MRF

A main difficulty in MR segmentation is the variation of the MR signal caused by inhomogeneities of the magnetic fields, especially of the excitation ($B_1$) field[3], and the sensitivity profile of the receiving coil. Therefore, a segmentation that is correct in one part of the volume will likely fail in another part.

This is corrected by adapting an additional, continuous inhomogeneity field **y**, also called the bias field [10]. As the magnetic field is slowly varying the *a priori* probability $p(\mathbf{y})$ must provide a smooth inhomogeneity field **y**. We present a new approach using a Markov random field with a local potential connecting 6 nearest neighbors:

---

[3] The tissue, the patient's head, may have significant impact on the inhomogeneity of the MR signal. In our approach we do not take into account this possible dependence, i.e. details on a possible origin of the inhomogeneity. Furthermore, the inhomogeneity is assumed to have a multiplicative effect on the MR echo intensities that is the same for all tissues.

$$p(\mathbf{y}) \propto e^{-U(\mathbf{y})} \qquad (4)$$

with the Gibbs energy

$$U(\mathbf{y}) = \alpha \sum_{<i,j>} (y_i - y_j)^2 + \beta \sum_i y_i^2 \,. \qquad (5)$$

At first sight it seems to be insufficient to regard a local neighborhood of 6 neighbors only to get a smooth inhomogeneity field. But it is well known from statistical mechanics that such local potentials yield high spatial correlations, as two distant points are connected by a number of local potentials[4]. Indeed, similar local potentials are used in physics to model ferromagnetic phases with infinite correlation length. The Gaussian MRF was chosen since the quadratic potential grows quickly and will therefore yield a smooth bias field. In image segmentation MRF with a linear potential for large neighbor differences are frequently used (e.g. see [17]). These potentials are of interest to preserve edges that are absent in the inhomogeneity field.

Further *a priori* information of the inhomogeneity field **y** is modeled by β, i.e. small inhomogeneity corrections are more probable than large. In section III we compare this approach to implementation of the inhomogeneity field used in the *adaptive segmentation* algorithm by Wells et al. [10]. While the theory of the *adaptive segmentation* is based on *a priori* bias field models, in the implementation the bias field is smoothed by a linear low-pass filter (eqn.(21) in [10]). This smoothing and the calculation of the segmentation are iterated.

Parzen-Window Intensity Distribution

Physical and chemical differences between tissues lead to different $T_1$ and $T_2$ Blochian relaxation times visualized in $T_1$ and $T_2$ weighted MR echos. Noise and in-tissue variations yield non-Gaussian, $T_1$-$T_2$ correlated distributions of the MR echo intensities.

---

[4] At n iterations the inhomogeneity field is influenced by all voxels in a nth order neighborhood.

Non-parametric statistics are used to describe the – insufficiently separated – tissue intensity distributions as accurately as possible. To this end the conditional probability of observing the echo intensity $z_i$ for given segmentation $x_i$ and inhomogeneity $y_i$ is modeled by a Parzen-window distribution [11]. For every tissue class $\bar{x}$ a set of $n_{\bar{x}}$ training points $\bar{z}_{k,\bar{x}}$ ($k=1,\ldots,n_{\bar{x}}$) is selected by a supervisor (see Kikinis et al. [23]). The Parzen-window distribution is obtained by centering a small Gaussian of width $\sqrt{\Sigma}$ around each training point

$$p(z_i | x_i, y_i) = \frac{1}{n_{x_i}} \sum_{k=1}^{n_{x_i}} \frac{1}{(2\pi)^{d/2} |\Sigma|^{1/2}} e^{-\frac{1}{2}(y_i z_i - \bar{z}_{k,x_i})^t \Sigma^{-1}(y_i z_i - \bar{z}_{k,x_i})} \quad (6)$$

where the covariance matrix $\Sigma$ is equal to $\sigma^2$ in the case of single-echo MR images (d=1) and to $\sigma^2$ times the two-dimensional unit matrix for double-echo MR images (d=2). In the latter case $y_i$, $z_i$, and $\bar{z}_{k,x_i}$ have two components, one for each echo.

This distribution deals with correlations between the echo signals. Even non-connected probability distributions can be described, e.g. SB tissue shows non-connected small *and* high echo intensities. For computational efficiency this probability distribution is stored in a look-up table as proposed in [23].

Simulated Annealing (SA)

A Markov process is used to generate random configurations {**x**,**y**} according to the probability distribution eqn. (1). A new configuration {**x'**,**y'**} is accepted with a probability W({**x**,**y**} → {**x'**,**y'**}) preserving detailed balance

$$p(\mathbf{x',y'}|\mathbf{z}) \, W(\{\mathbf{x',y'}\} \to \{\mathbf{x,y}\}) =$$
$$p(\mathbf{x,y}|\mathbf{z}) \, W(\{\mathbf{x,y}\} \to \{\mathbf{x',y'}\}). \quad (7)$$

According to Metropolis et al. [14] a prototype acceptance probability is

$$W(\{\mathbf{x},\mathbf{y}\} \to \{\mathbf{x'},\mathbf{y'}\}) = \min\left\{1, \frac{p(\mathbf{x'},\mathbf{y'}|\mathbf{z})}{p(\mathbf{x},\mathbf{y}|\mathbf{z})}\right\}. \tag{8}$$

Due to the local potential chosen, this probability can be calculated effectively when (**x'**,**y'**) differs from (**x**,**y**) at a single voxel. Since detailed balance is fulfilled and every configuration is accessible from any other configuration, ergodicity is established [12] and the Markov process will yield segmentations **x** and inhomogeneities **y** obeying eqn.(1). Testing one new segmentation and inhomogeneity for every voxel is called a Monte-Carlo sweep. From a set of sweeps one may calculate mean values as well as standard deviations and correlations within the segmentation. We follow a slightly different approach to obtain the MAP estimate instead of these mean values, i.e. simulated annealing.

Simulated annealing was introduced by Kirkpatrick et al. [24] and refers to the slow decrease of a control parameter T that corresponds to temperature in physical systems. We modify eqn.(1) in the following way

$$p(\mathbf{x},\mathbf{y}|\mathbf{z}) \to \exp\left(\frac{1}{T}\log(p(\mathbf{x},\mathbf{y}|\mathbf{z}))\right) \tag{9}$$

As T decreases, samples from the *a posteriori* distribution are forced towards the minimal energy, i.e. towards the MAP. If T(l) satisfies

$$T(l) \geq \frac{\text{const}}{\log(1+l)} \tag{10}$$

at the lth Monte-Carlo sweep, it can be proved that this annealing schedule theoretically guarantees convergence to the global MAP [12]. The starting configuration for the SA algorithm is zero inhomogeneity **y**=0 and the segmentation **x,** that would be chosen by a point-by-point segmentation solely due to the Parzen-window distribution, maximizing eqn.(6) with **y**=0 (this is equivalent to eqn.(14) if the neighborhood information, not available at this time, is disregarded). The temperature was reduced according to T(l)=1/log(1+l).

Iterated Conditional Modes (ICM)

Especially for clinical applications the 3D SA-segmentation algorithm requires too much computing time, at least for current single processor systems. Therefore, we present an alternative method, i.e. *iterated conditional modes* proposed by Besag [21]. Following this concept, the algorithm maximizes the *a posteriori* probability with respect to the segmentation **x** and the inhomogeneity **y** iteratively.

In a first step for every voxel *i*, the most probable discrete segmentation $x_i$, maximizing eqn.(1) at fixed **y** and neighboring segmentation $\mathbf{x}_{\partial_i}$ is chosen[5]:

$$\max_{x_i}\{p(\mathbf{x},\mathbf{y}|\mathbf{z})\} = \qquad (11)$$

$$\max_{x_i}\left\{p(z_i|x_i,y_i)e^{-\sum_{j\in\partial_i}e_{x_ix_j}}\right\}.$$

In a second step the continuous inhomogeneity $y_i$ is maximized for every voxel *i* at fixed **x** and neighboring inhomogeneity $\mathbf{y}_{\partial_i}$

$$\frac{\partial}{\partial y_i}p(\mathbf{x},\mathbf{y}|\mathbf{z}) = 0. \qquad (12)$$

To solve this equation with respect to $y_i$ the Parzen-window distribution (6) is too complicated. Therefore, we follow Wells et al. [10] and heuristically replace eqn.(6) by a Gaussian distribution for the logarithmic echo intensities $\tilde{z}_i = \ln z_i$ with tissue dependent mean $\bar{z}_i$ and covariance matrix $\Sigma_{x_i}$

---

[5] If the maximal probability is below some fixed value the voxel is labeled as unclassified. Unclassified voxels are indifferent in the potential $e_{x_ix_j}$.

$$p(z_i \mid x_i, y_i) = \frac{1}{(2\pi)^{d/2} |\Sigma_{x_i}|^{1/2}} e^{-\frac{1}{2}(\tilde{z}_i + y_i - \bar{z}_{x_i})^t \Sigma_{x_i}^{-1}(\tilde{z}_i + y_i - \bar{z}_{x_i})}. \quad (13)$$

Differentiation of eqn.(12) yields

$$\frac{\partial}{\partial y_i}\left\{\frac{1}{2}(\tilde{z}_i + y_i - \bar{z}_{x_i})^t \Sigma^{-1}(\tilde{z}_i + y_i - \bar{z}_{x_i}) + \alpha \sum_{j \in \partial_i}(y_i - y_j)^2 + \beta y_i^2\right\} = 0.$$

Solving this equation gives the most probable inhomogeneity $y_i$ for fixed segmentation and neighborhood inhomogeneities:

$$y_i = A^{-1}\left\{2\alpha \Sigma_{x_i} \sum_{j \in \partial_i} y_j + (\bar{z}_{x_i} - \tilde{z}_i)\right\} \quad (14)$$

with

$$A = (2\alpha n + 2\beta)\Sigma_{x_i} + 1, \qquad n = 6.$$

Following eqn.s (11) and (14) an iterated conditional modes algorithm (ICM1) is implemented. The algorithm ICM1 begins, as SA, with zero inhomogeneity **y**=0 and the segmentation **x** maximizing eqn.(6) for **y**=0. From this segmentation an inhomogeneity field is calculated according to eqn.(14). The inhomogeneity field on the other hand, together with the old neighborhood segmentation, allows determination of the new segmentation according to eqn.(11). Continuing the iteration of eqn.s (14) and (11) will converge to some local, not necessary global maximum of the *a posteriori* probability.

Another algorithm (ICM2) using the inhomogeneity approach of Wells et al. [10] is presented.

Here step (14) is replaced by calculating the inhomogeneity $y_i$ locally

$$y_i = \frac{R_i}{N_i} \quad (15)$$

with residual

$$R_i = \sum_{x_i} p(z_i | x_i, y_{i_{old}}) p(x_i) \Sigma_{x_i}^{-1} (\bar{z}_{x_i} - \tilde{z}_i) \qquad (16)$$

and normalizer

$$N_i = \sum_{x_i} p(z_i | x_i, y_{i_{old}}) p(x_i) \Sigma_{x_i}^{-1}. \qquad (17)$$

In addition, all possible segmentations $x_i$ are evaluated according to their probability $p(z_i|x_i, y_{i_{old}})p(x_i)$ at the old inhomogeneity field $y_{i_{old}}$. Instead of the *a priori* inhomogeneity distribution (4) a linear low-pass filter is used to obtain a smooth residual $R_i$ and normalizer $N_i$ before calculating the inhomogeneity **y** according to eqn.(15). Further details can be found in section 2.3 of [10]. ICM2 is initialized in the same way as ICM1 and SA, and then iterates eqn.s (15) and (11).

Parameter Choice

The strength of the MRF is that simple modeling of this type yields additional information that considerably helps to segment otherwise ambiguous voxels. The MRF probability should not been oversized since a cluster of voxels is first of all classified according to its MR signal. Using faithful, anatomical potentials $e_{x_i x_j}$ that were calculated from the statistics of a segmented image does not always improve the segmentation since some probabilities were extremely low, e.g. to have a CSF voxel in a GM/WM neighborhood. Consequently, in parts of the image, CSF voxels were misclassified as WM/GM. This led us to the following considerations for setting suitable parameters.

To take all three features into account (i.e. neighborhood correlations, signal inhomogeneities and Parzen-window distribution of tissue intensities), the energies (3), (5), and (6) must have the same order of magnitude. As the tissue intensities typically differ from their mean value by the standard deviation, the corresponding energy is 1/2 for eqn.(6). The neighborhood contribution (3) must have the same order of magnitude. Two different energies $e_{x_i x_j}$ for neighboring voxels of the same (set to 0) and of different tissues (set to $\varepsilon$) are used. With 6 neighbors contributing $\varepsilon$ should be about $1/2 \times 1/6$. As the contribution due to the MR intensity is more important than this *a priori* knowledge, we actually choose $\varepsilon = 0.05$.

The *a priori* information on the inhomogeneity is described by eqn. (5). The parameter $\beta$ corresponds to the expected inverse variance of the inhomogeneity field, e.g. if the inhomogeneities have a standard deviation of $\Delta I = 0.1$ then $\beta = 1/(2\,\Delta I^2) = 50$. In order to allow larger inhomogeneities and not to be too severly restricted by this *a priori* information we used $\beta = 20$. The parameter $\alpha$ describes the standard deviation of the inhomogeneity gradient. To obtain a smooth inhomogeneity field $\alpha$ is set to 100 allowing an inhomogeneity variation of $I = 0.1$ within about 5 voxels.

## B. Acquisition of MR Images

Brain MR images of six volunteers were acquired on a Siemens Vision, 1.5 Tesla. The sequence used was a turbo spin echo $pd/T_2$ weighted double-echo with $T_R$=6536 ms, $T_{E1}$=20 ms and $T_{E2}$=120 ms. At an image resolution of 1x1x1 mm$^3$ a field of view of 256x160x32 mm$^3$ (32 slices for each dataset) was scanned within 10.5 min. For this sequence the equipment specific noise, as estimated by the mean BG signal, was about 30 (relative units) for both echos. Mean tissue signals (standard deviations) were calculated from a segmented MR slice:

pd-echo:   SB=456(120),  WM=823(70),  GM=1059( 95),  CSF=1363(177);

$T_2$-echo:   SB=167( 69),  WM=426(59),  GM= 602(102),  CSF=1223(307).

These two echos are correlated and a criterion that takes this into account is the Mahalanobis distance D between two tissues ($\bar{x}$ and $\bar{x}'$), i.e. $D^2 = (\bar{z}_{\bar{x}} - \bar{z}_{\bar{x}'})^T \Sigma^{-1} (\bar{z}_{\bar{x}} - \bar{z}_{\bar{x}'})$ where $\Sigma$ is the mean value of the two covariance matrices of the two tissues. As one can already guess from the mean echo signals and their standard deviations, the tissues that are most difficult to distinguish are WM-GM (D=2.9) and GM-CSF (D=3.0).

For a direct qualitative comparison, the single-echo algorithm was tested using the *pd* weighted image which has the better WM-GM contrast of the two MR echos.

### C. Simulation of MR Images

To validate the segmentation algorithms as well as to compare them quantitatively we have used simulated MR images. A template segmentation of a 256x256x16 MR image with 1 mm isotropic resolution was used. Smoothing, noise and inhomogeneity were modeled as follows (see Fig. 1):

1. The template segmentation was defined as the correct segmentation. From the corresponding double-echo MR image (see II-B) the mean values were calculated for all tissues. Each voxel intensity was set as specified in Section II.B.

2. This image was smoothed with 6 nearest, first order, neighbors in 3D. The weight of every neighboring voxel relative to the weight of the center voxel was denoted as *S*. At a value of *S*=1/6 the echo intensity of one voxel was 50% due to its tissue and 50% due to the neighboring tissues.

3. 2D-Gaussian noise of standard deviation *N* was added. A relative measure for this noise was the contrast to noise ratio between two tissues ($\bar{x}$ and $\bar{x}'$): $CNR_{\bar{x},\bar{x}'} = (\bar{z}_{\bar{x}'} - \bar{z}_{\bar{x}})/N$. At the above mean *pd* echo signals a typical noise level of $N = 50$ corresponded to a contrast-to-noise ratio $CNR_{WM,GM} = 4.7$ and to a signal to noise ratio $SNR_{GM} = 21$ for GM.

4. The image was multiplied by an inhomogeneity factor which changed linearly with Euclidean distance from the point $(x,y,z) = (0,0,15/2)$. The maximal inhomogeneity factor at a non-background tissue was set to $1 + I$, whereas the minimal factor was set to $1 − I$. In this way $I$ was a measure of the strength of the inhomogeneity. At $I = 0.125$ the local GM *pd* echo mean varied from 927 in one part of the simulated image to 1191 in another. The CSF variation was between 1192 and 1533. Therefore at $I = 0.125$ the inhomogeneity was so large that CSF had the same mean intensity in one part of the image as GM in the other.

## III. RESULTS

We simulated MR images with different noise ($N$), inhomogeneity ($I$) and smoothing ($S$). These images were segmented and compared to the correct segmentation, i.e. the one the simulation was based on. The error was calculated as the ratio of mis-classified voxels to non-background voxels within the entire 256x256x16 volume.

First we analyzed the convergence of the algorithms at a typical noise level $N = 50$, inhomogeneity $I = 0.1$ and smoothing $S = 0.2$. In Fig. 2 no SB tissue was simulated. It shows that at 6 iterations an adequate convergence was obtained, only the ICM1 algorithm slightly improved with further iterations. Therefore further comparisons were done at 6 iterations. In Fig. 3 the SB tissue was simulated additionally. In this case the ICM2 and the *adaptive segmentation* (AS) algorithms had huge errors of more than the number of brain voxels after convergence. The reason for this was that many background voxels were misclassified as SB. In practice one can reduce these errors by pre- and postfiltering or by attaching a low *a priori* probability for SB. For an appropriate comparison we did not further simulate the SB structure.

The impact of noise was analyzed in Fig. 4. Up to $N = 50$ the ICM1 and ICM2 algorithms show low error rates of less than 0.5% whereas the *adaptive segmentation* algorithm misclassified 2.3% of the voxels. For noise levels of $N = 30\text{-}100$ the single-echo ICM1 yielded better results than the *adaptive segmentation* algorithm. For all algorithms the error increased exponentially with the standard deviation $N$ of the Gaussian noise.

The influence of inhomogeneity was simulated by different values of $I$, see Fig. 5. For the whole range of tested inhomogeneities the *adaptive segmentation* algorithm yielded an almost constant error of 2.3%. Whereas at inhomogeneities $I < 0.10$ both ICM algorithms performed better than the *adaptive segmentation* algorithm, at inhomogeneities larger than $I = 0.10$, i.e. larger than the tissue contrast, the ICM1 algorithm performed worse.

Smoothing, controlled by the parameter S, wipes out fine details of the simulated MR image. Fig. 6 shows that all algorithms had difficulty to reconstruct these details, the error considerably grew with S. Thereby, ICM1 performed slightly better with smoothing.

To analyze the influence of tissue thickness on the segmentation error we did not use the brain template but an artificial image with a sinusoidal GM structure between WM and BG. The GM „gyrus" was $d$ voxels thick. The error was calculated as misclassified voxels in a 11 voxel neighborhood of the GM structure divided by the total number of GM voxels. At $N = 50$, $I = 0.1$ and $S = 0.2$ the ICM algorithms yielded good results (see Fig. 7). The error did not improve significantly for $d > 3$, whereas it was moderately higher for a fine GM structure of one voxel thickness ($d=1$).

Fig. 8 compares the *simulated annealing* (SA) to the ICM1 implementation for two different sets of parameters. The SB tissue was simulated again. It was demonstrated, that *simulated annealing* reduced the error, e.g. by 4% for $N = 50$, $I = 0.1$, and $S = 0.2$. This reduction was smaller if the SB tissue was not simulated.

For real brain MR images only a qualitative comparison of the algorithms was possible since the correct segmentation is unknown and therefore a quantitative comparison of the two algorithms did not reveal which one was wrong. Consequently, we show only the segmentations obtained and qualitatively point out some obvious failures. Fig. 9 displays one slice of a 3D double-echo MR image. Fig. 10 shows the corresponding segmentations. For this slice the *adaptive segmentation* algorithm performed poorly at 6 *iterations* misclassifying many WM voxels as GM. Therefore, the result for the first *iteration* is displayed too. At one *iteration* no inhomogeneity field was adapted and the segmentation was slightly worse in the occipital part of the brain (see bottom of Fig.10b). The MRF neighborhood correlations improved the segmentation of the non-brain scalp-bone structure and reduced single-voxel misclassifications. Based on the *pd* weighted image the single-echo ICM1 and *simulated annealing* (SA) algorithms were able to classify the MR brain into GM, WM, CSF, and SB.

## IV. Discussion

We presented three new segmentation algorithms (ICM1, ICM2, and SA) that incorporate non-parametric distributions of tissue intensities, and model neighborhood correlations as well as signal inhomogeneities within the MR image. They differ in the method to handle the inhomogeneity field (ICM2 vs. ICM1/SA) and to maximize the *a posteriori* probability (SA vs. ICM1/ICM2).

The ICM2 algorithm captures the intensity distribution and the inhomogeneity field in the same way as the *adaptive segmentation* [10] algorithm does, but ICM2 models neighborhood correlations as additional information to improve the segmentation The main improvement of the use of the MRF is its better noise resistance. This can be seen in Fig. , where ICM2 shows an error rate of less than 0.5% up to CNR=2.4. Smoothing and even inhomogeneities are better captured (see Fig. 5 and Fig. 6).

The heuristic smoothing of the inhomogeneity field, as do the *adaptive segmentation* and ICM2 algorithms, does not take into account that small inhomogeneity corrections are more probable than large ones. Therefore, the ICM2 and the AS algorithms have the tendency to overestimate the inhomogeneity. Upon more iterations the segmentation becomes worse, see Fig. 2 and especially Fig. 3 where the additional simulation of scalp-bone structure drastically increased the error. In practice, this has the consequence that the operator often has to try different numbers of iterations to yield the best segmentation result, see Fig. 10. Additionally, at fewer iterations the inhomogeneities at the outer slices are not captured correctly. The ICM1 as well as the SA algorithm achieved further improvement by the use of an *a priori* Gibbs distribution for the inhomogeneity field. The ICM2 algorithm is superior to ICM1, see Fig. 4, only at inhomogeneities larger than the tissue contrast which were not observed in our MR images.

The large error of the *adaptive segmentation* algorithm for the artificial simulated sinusoidal image to test the GM thickness (Fig. 7) is based on the fact that at small GM structures the smoothing lowers the GM intensity. Therefore, many GM voxels were misclassified as WM when using the *adaptive segmentation* algorithm. At large GM structures the inhomogeneity was not captured correctly leading to a great error.

The *simulated annealing* algorithm (see Fig. 8) improves the segmentation, with the penalty being long computing times. One needs 1000 SA sweeps instead of 10 ICM iterations to obtain accurate results[6]. This shortcoming can be overcome by parallelizing the SA algorithm. Herein lies an important advantage of the local structure of the MRF: test and update of new configurations are possible at the same time and are only based on the information within a small neighborhood of a voxel.

## V. CONCLUSION

We have presented a *simulated annealing* and an *iterated conditional modes* version of a new Markov random field 3D segmentation algorithm. After the training of typical echo intensities and setting one MRF parameter ($\beta$) according to the expected inhomogeneity (see p.14) the algorithm fully automatically segments the entire 3D MR volume as well as different MR images, that are acquired with the same MR sequence. For the first time non-parametric intensity distributions, neighborhood correlations, and inhomogeneities are combined in one segmentation algorithm. Volunteer MR images and detailed, quantitative simulations demonstrated that only the combination of these three features leads to an accurate and robust segmentation with respect to noise, inhomogeneities, and structure thickness. *Simulated annealing* allows an additional improvement compared to *iterated conditional modes* at the expense of long running times.

---

[6] This corresponds to 2000 minutes instead of about 20 minutes for the segmentation of a 256x256x64 MR volume on a SPARC20 workstation.

*Acknowledgment*
We thank S. Posse, H. Herzog and L. Tellmann for their advice and technical assistance. We appreciate the help of Dr. J. Shah in preparing the manuscript. The cooperation between the Forschungszentrum Jülich, Institute of Medicine and the Harvard Medical School, Brigham & Women's Hospital was supported by the DFG grant Ro 1149/1-1; the work in the Harvard Medical School, Brigham & Women's Hospital was supported by the Whitaker Foundation grant, by the NIH pO1 CA 67165-01A1 grant, and by the NSF BES-9631710 grant.

FIGURE CAPTIONS

Fig. 1. The four tasks performed to simulate double-echo (pd (left) and $T_2$ (right)) MR images: The intensities are initialized according to their mean intensities (a), afterwards smoothed (b; S=0.2), noised (c; N=50), and finally an inhomogeneity is simulated (d; I=0.1).

Fig. 2. Convergence of the segmentation algorithms without the simulation of the scalp-bone tissue. A noise level of *N*=50, additional inhomogeneity *I*=0.1 and smoothing *S*=0.2 is simulated. Plotting the error rate vs. number of iterations shows that at 6 *iterations* the convergence is adequate. (ICM=*iterated conditional mode*; AS=*adaptive segmentation*; SE=*single-echo*; DE=*double-echo*)

Fig. 3. Convergence of the segmentation algorithms with scalp-bone tissue. A noise level of *N*=50, additional inhomogeneity *I*=0.1 and smoothing *S*=0.2 is simulated. The ICM2 and the *adaptive segmentation* algorithm fail to correct the inhomogeneity.

Fig. 4. Error rate vs. noise level at *I*=0, *S*=0 and 6 iterations. The error rate increases exponentially with the Gaussian noise *N*. The ICM algorithms show no significant error up to *N*=50 or $CNR_{WM,GM}$=4.7.

Fig. 5. Influence of inhomogeneity *I* at fixed *N*=50, *S*=0 and 6 iterations. Up to *I*=0.10 all algorithms correct the inhomogeneity well. For higher inhomogeneities the ICM1 algorithm performs worse.

Fig. 6. Error rate vs. smoothing at $N$=50, $I$=0 and 6 iterations. The information loss due to the smoothing S cannot be recovered and thus the error rate increased drastically.

Fig. 7. Error rate vs. GM thickness $d$. When using Markov random fields fine structures can be resolved at a low error that is only somewhat enhanced for fine GM structures of one voxel thickness ($d$=1).

Fig. 8. *Simulated annealing* (SA) compared to ICM1 for single-echo simulations with scalp-bone structure at two parameter sets (A: $N$=50, $I$=0.1, $S$=0.2; B: $N$=80, $I$=0, $S$=0).

Fig. 9. Double-echo, $pd$ (left) and $T_2$ (right) weighted transversal slices of a 3D brain MR image.

Fig. 10. Segmentations of the MR images from Fig. 9 (*white*: background, *pink*: scalp-bone, *yellow*: csf, *red*: grey matter, *blue*: white matter, and *brown*: unclassified) as determined by the AS algorithm at 6 and 1 iteration ( a and b, respectively), double-echo ICM2 (c) and ICM1 (d), and finally by the single-echo versions of ICM1 (e) and SA (f).

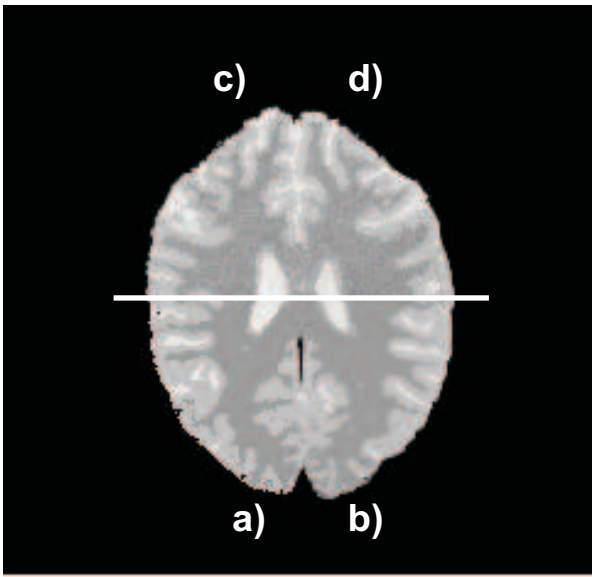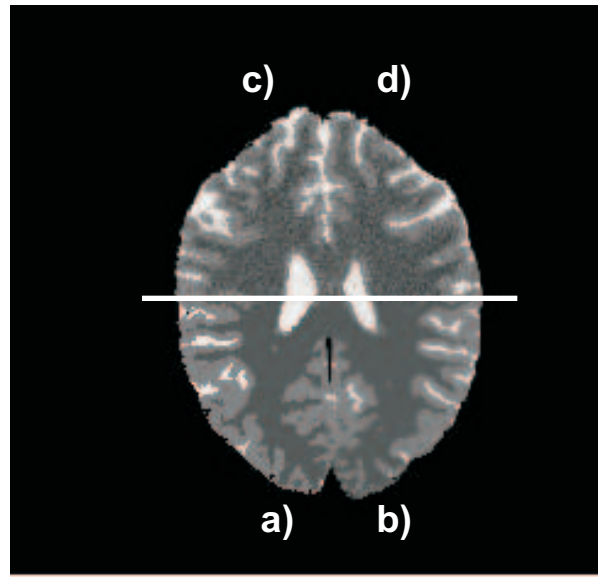

**Fig. 1**

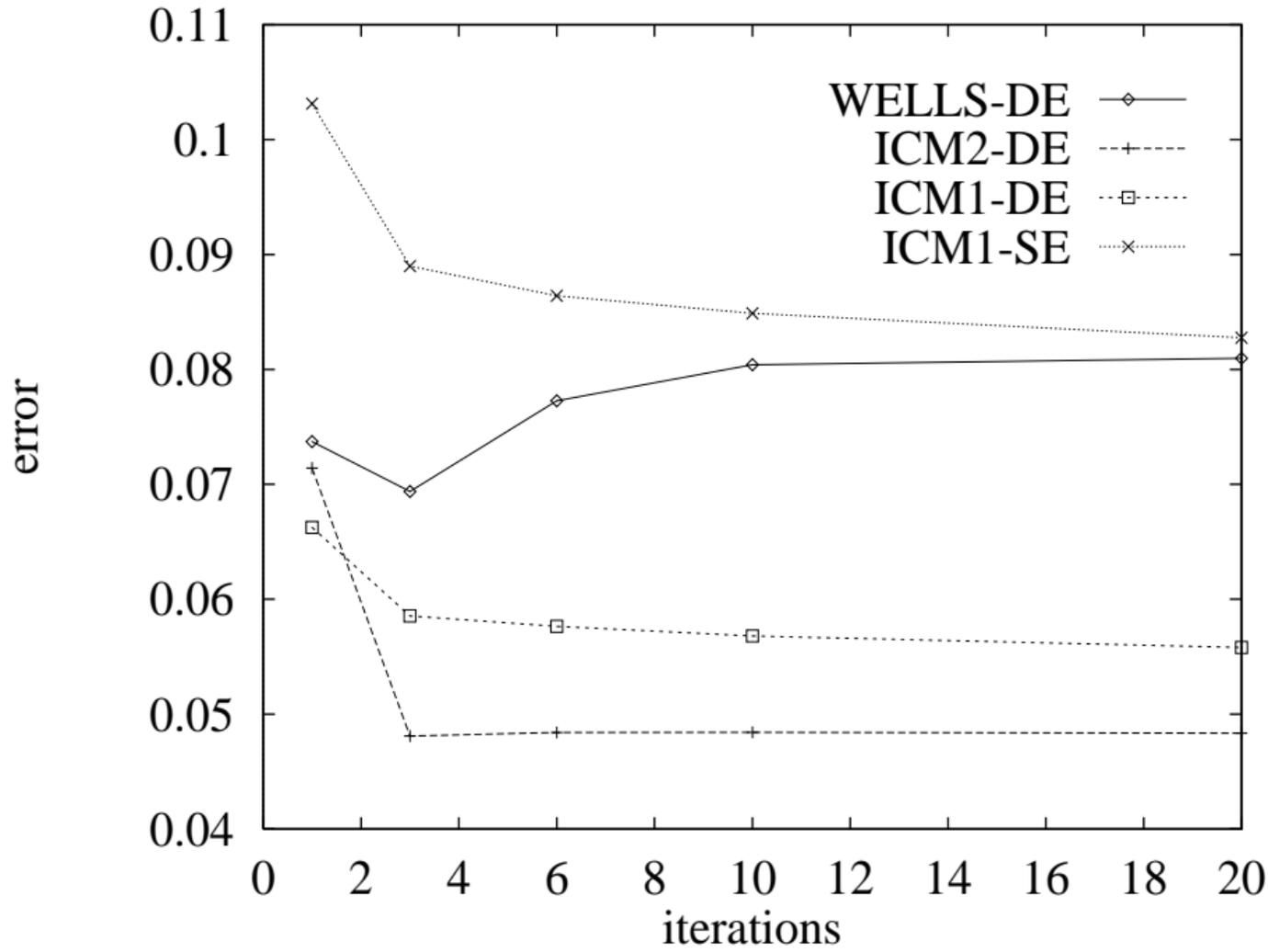

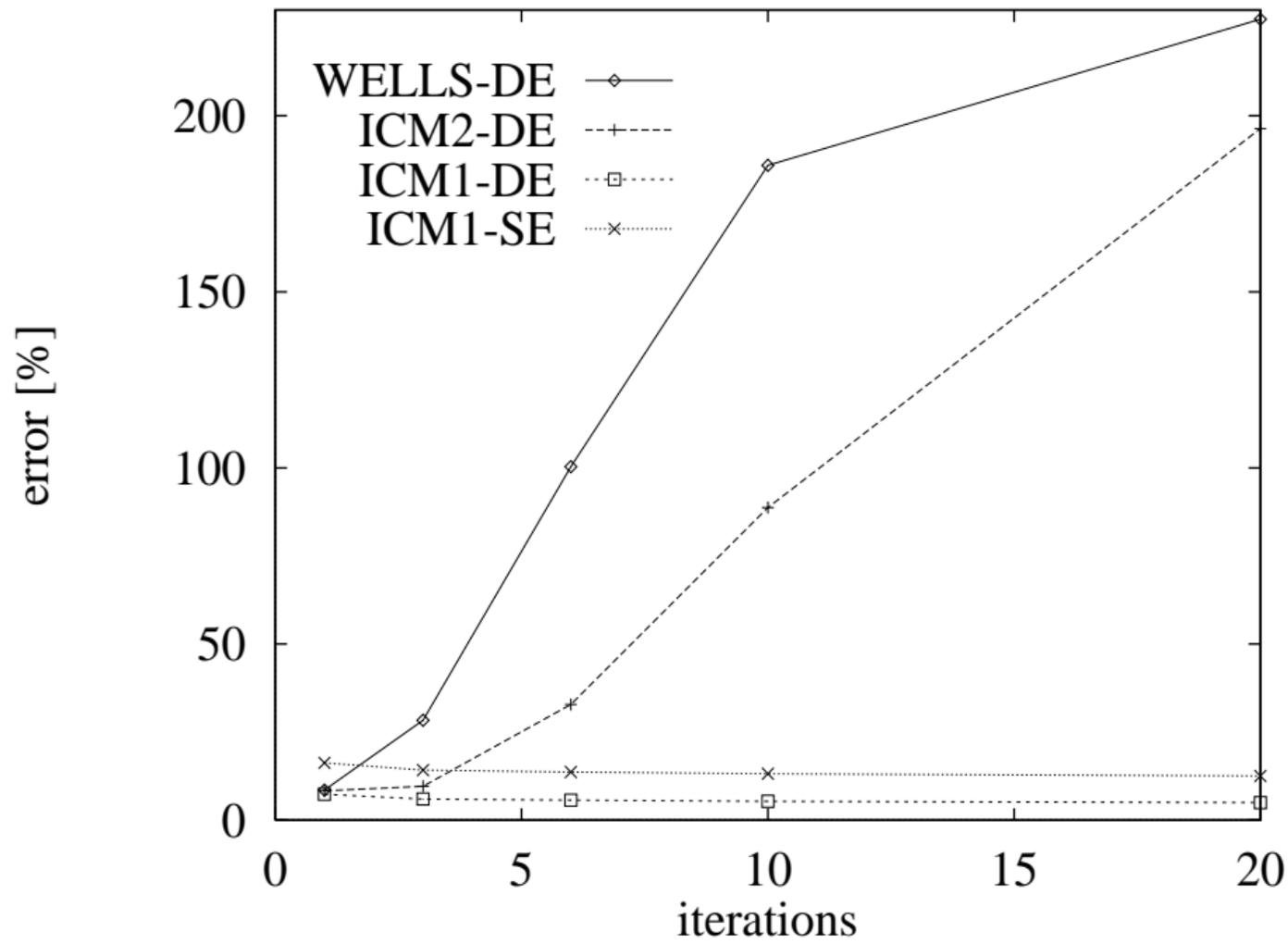

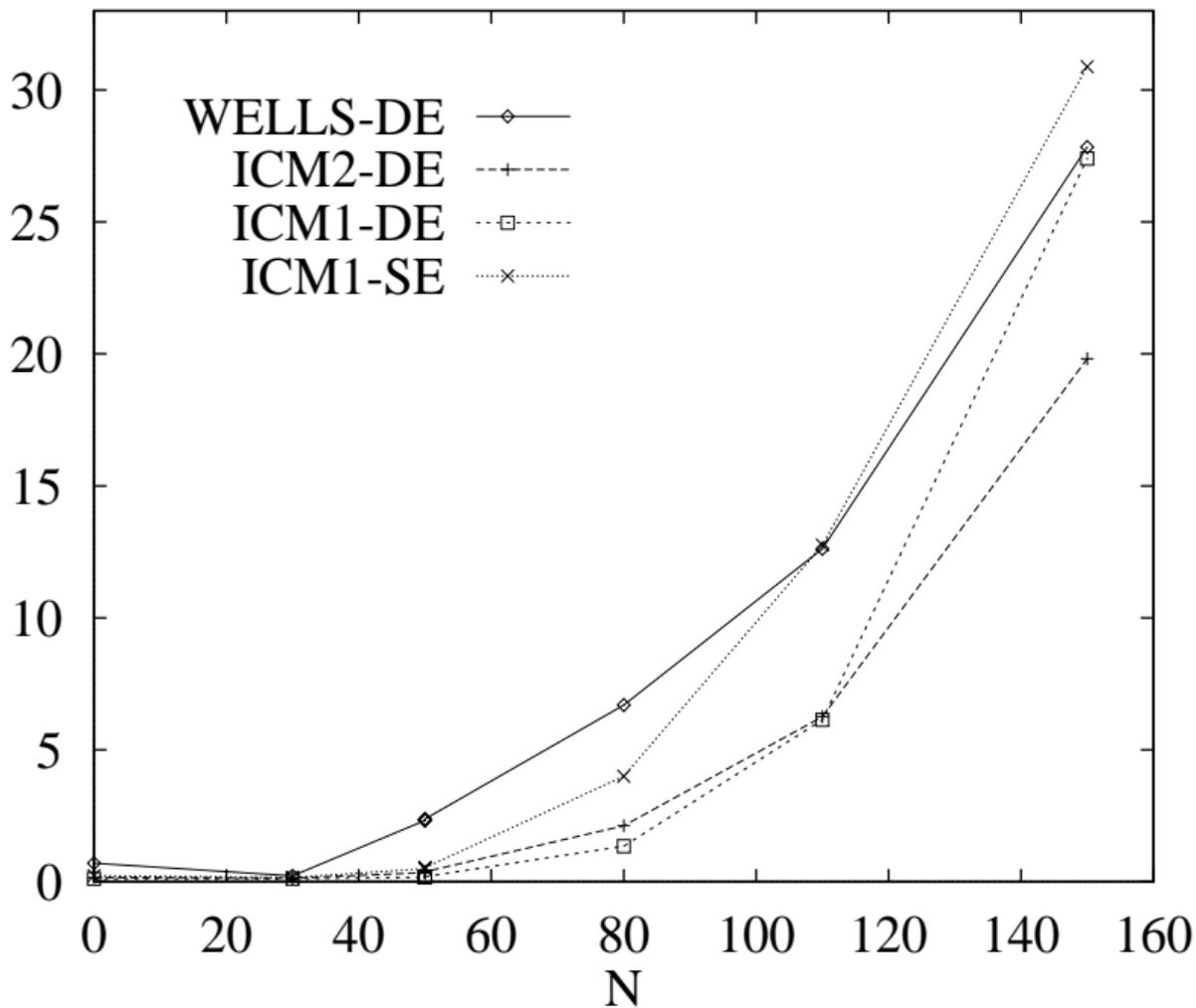

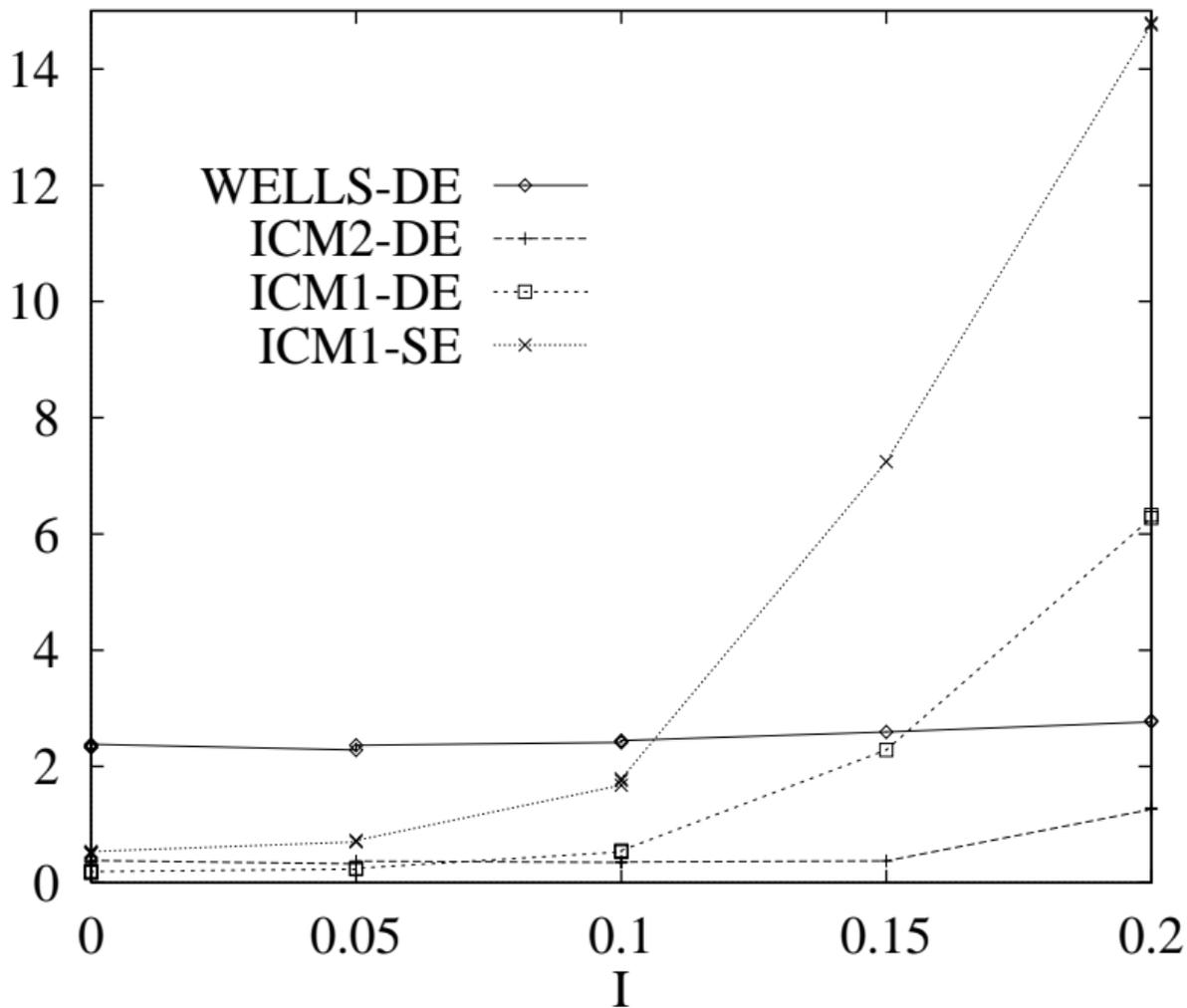

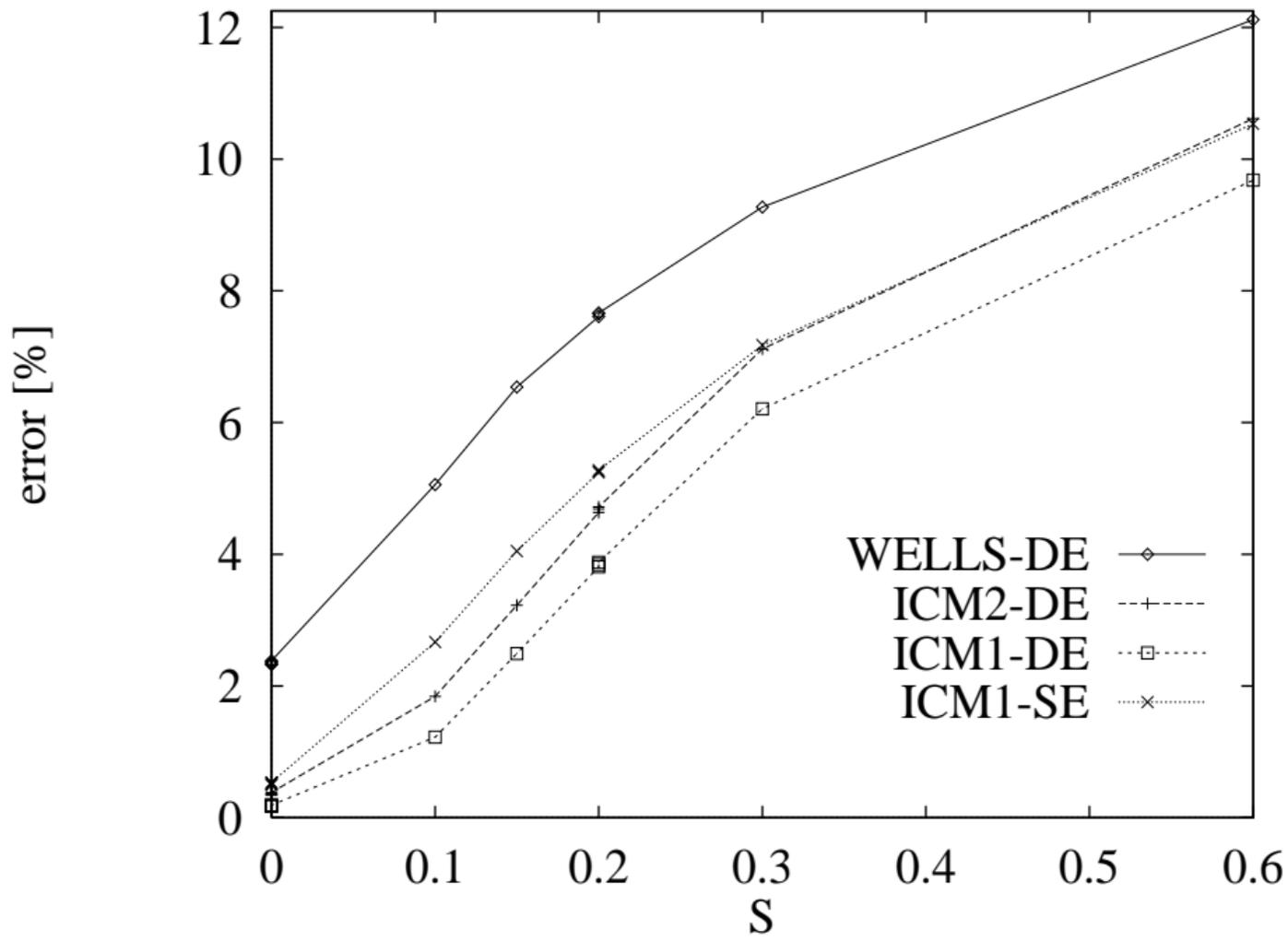

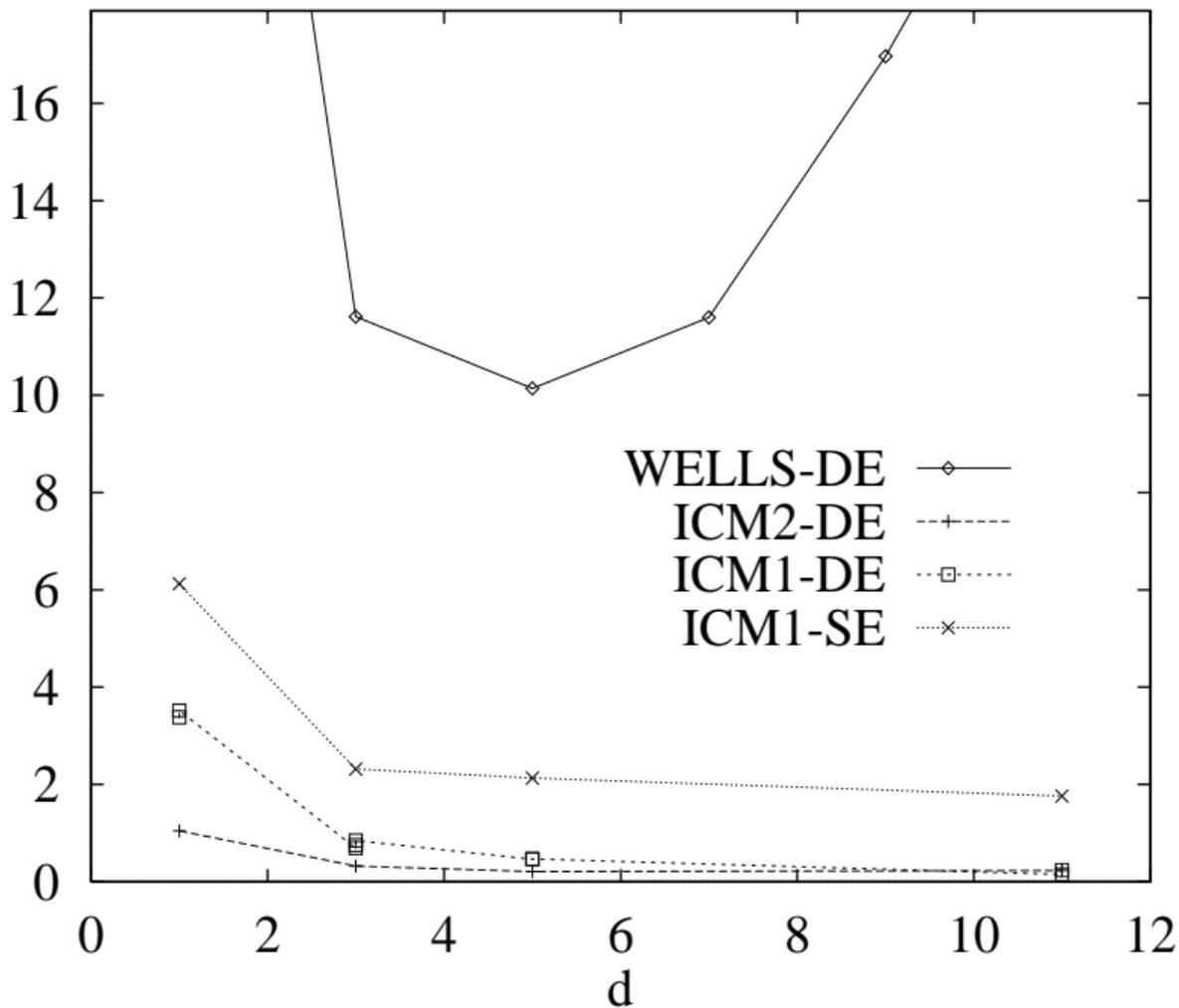

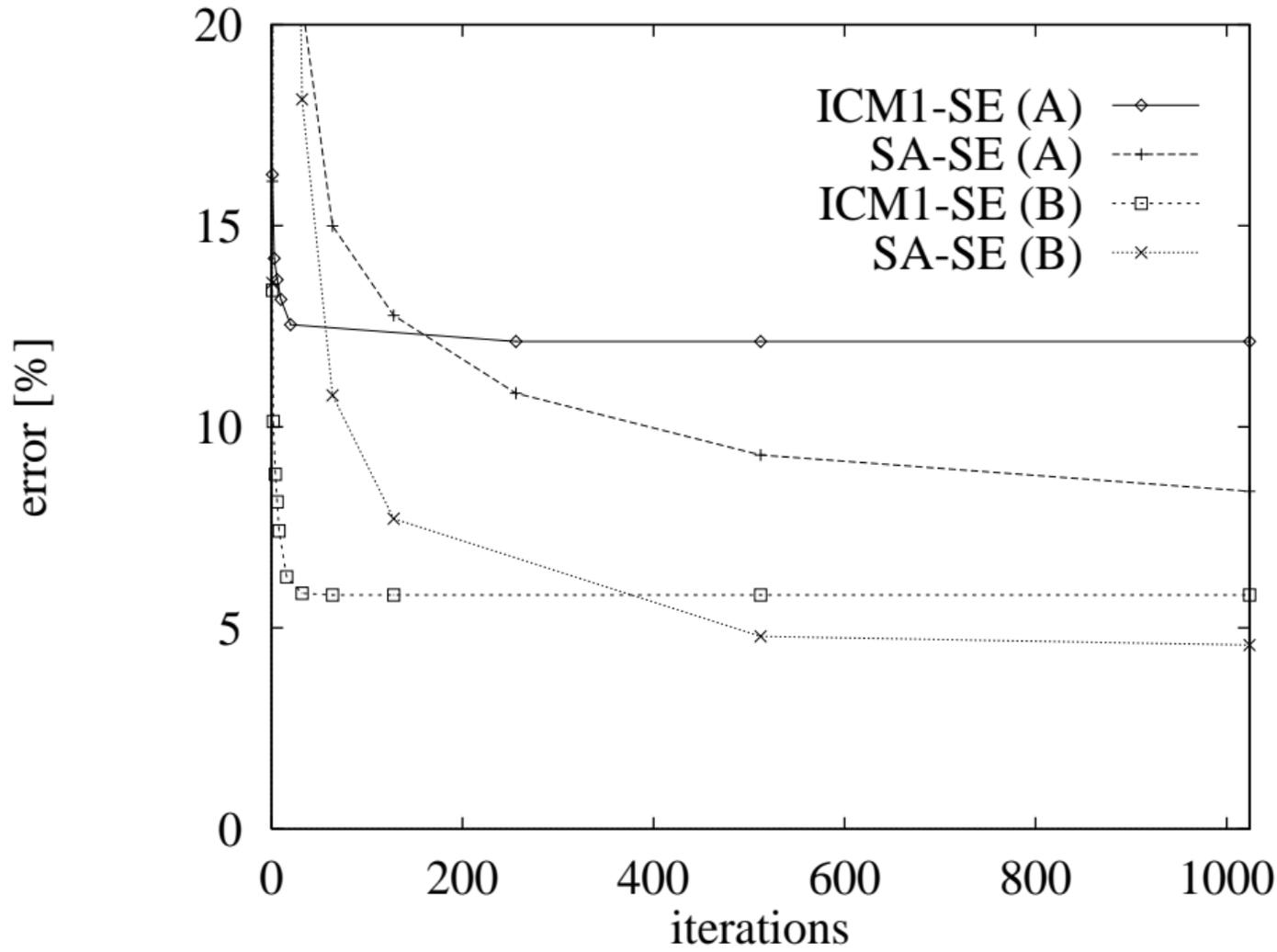

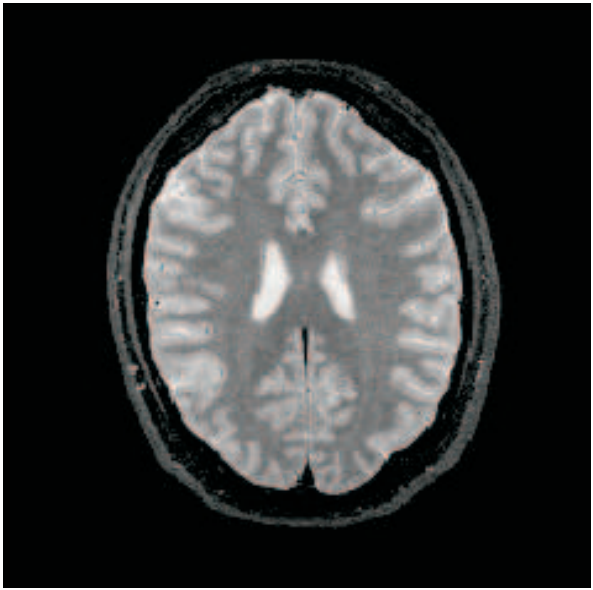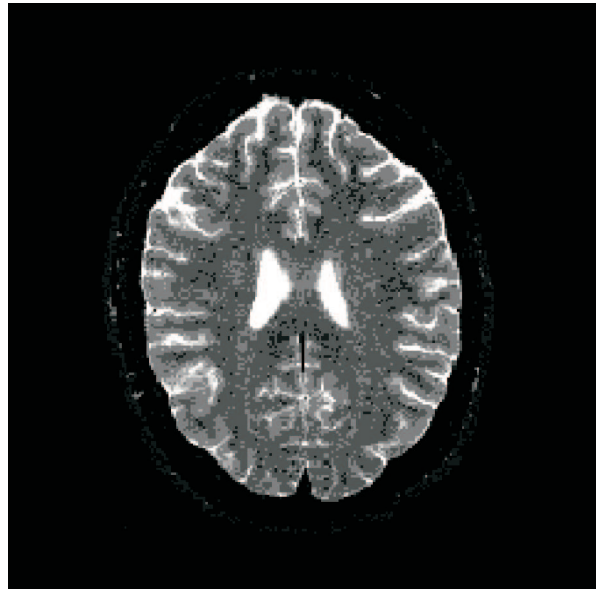

**Fig. 9**

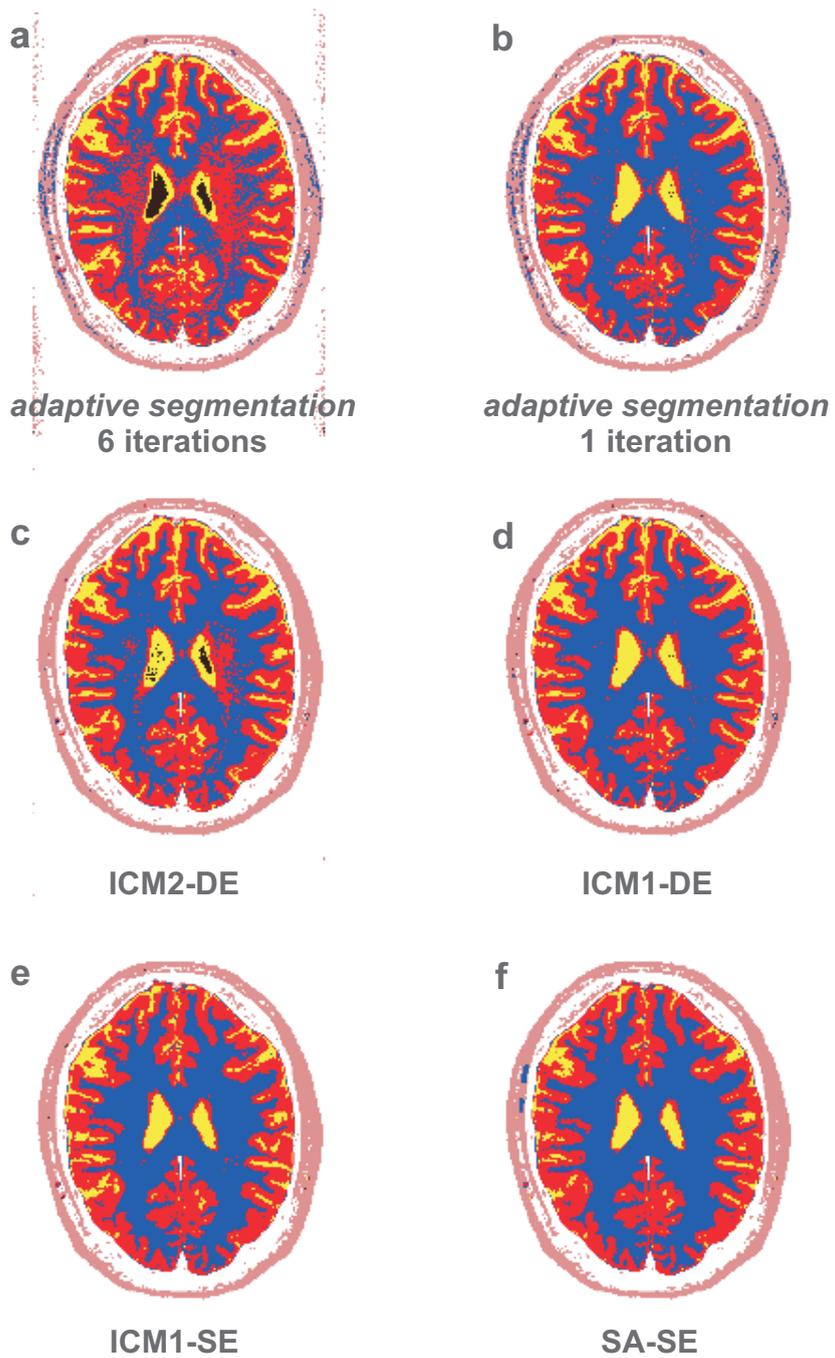

**Fig. 10**